\documentclass[]{opendatalab}
\usepackage{xcolor}
\usepackage{longtable}
\usepackage{listings}
\usepackage[numbers]{natbib}

\usepackage{adjustbox}
\usepackage{multirow}
\usepackage{multicol}
\usepackage{booktabs}
\usepackage{setspace}
\usepackage{hyperref}
\usepackage{threeparttable}
\usepackage{colortbl}
\usepackage{float}
\usepackage{enumitem}
\usepackage{array}
\usepackage{arydshln}
\usepackage{makecell}
\usepackage{amsmath}
\usepackage{dashrule}
\usepackage{xspace}
\newcommand{\method}{ScaleDiff\xspace}
\newcommand{\gen}{DiffGen-8B\xspace}

\definecolor{codegreen}{rgb}{0,0.6,0}
\definecolor{codegray}{rgb}{0.5,0.5,0.5}
\definecolor{codepurple}{rgb}{0.58,0,0.82}
\definecolor{backcolour}{rgb}{0.95,0.95,0.92}
\definecolor{promptcolor}{HTML}{D1D0F2}
\definecolor{promptcolorheader}{HTML}{bdbcec}
\newcommand{\promptbox}[2]{
\begin{tcolorbox}[
top=0.3em,bottom=0.3em,left=0.5em,right=0.5em,
toptitle=0.3em,bottomtitle=0.2em,boxsep=0pt,
colframe=promptcolorheader,colback=promptcolor!50,boxrule=0.5pt,
]
\footnotesize
\end{tcolorbox}
}

\newlength\savewidth\newcommand\shline{\noalign{\global\savewidth\arrayrulewidth
        \global\arrayrulewidth .8pt}\hline\noalign{\global\arrayrulewidth\savewidth}}

\lstdefinestyle{mystyle}{
    backgroundcolor=\color{backcolour},   
    commentstyle=\color{codegreen},
    keywordstyle=\color{magenta},
    numberstyle=\tiny\color{codegray},
    stringstyle=\color{codepurple},
    basicstyle=\ttfamily\footnotesize,
    breakatwhitespace=false,         
    breaklines=true,                 
    captionpos=b,                    
    keepspaces=true,                 
    numbers=left,                    
    numbersep=5pt,                   
    showspaces=false,                
    showstringspaces=false,
    showtabs=false,                  
    tabsize=2
}

\title{ScaleDiff: Scaling Difficult Problems for Advanced Mathematical Reasoning}

\author{Qizhi Pei$^{1,2,\dagger,\ddagger}$ ~~Zhuoshi Pan$^{2,3,\dagger,\ddagger}$ ~~Honglin Lin$^{2}$ ~~Xin Gao$^{2}$ ~~Yu Li$^{2}$\\ \textbf{Zinan Tang$^{2}$}
~~\textbf{Conghui He$^{2,*}$} ~~\textbf{Rui Yan$^{2,4,*}$} ~~\textbf{Lijun Wu$^{2,*}$}
}

\affiliation[1]{Gaoling School of Artificial Intelligence, Renmin University of China}
\affiliation[2]{OpenDataLab, Shanghai Artificial Intelligence Laboratory}
\affiliation[3]{Tsinghua University}
\affiliation[4]{School of Artificial Intelligence, Wuhan University}

\abstract{
Large Reasoning Models (LRMs) have shown impressive capabilities in complex problem-solving, often benefiting from training on difficult mathematical problems that stimulate intricate reasoning. 
Recent efforts have explored automated synthesis of mathematical problems by prompting proprietary models or large-scale open-source models from seed data or inherent mathematical concepts.
However, scaling up these methods remains challenging due to their high computational/API cost, complexity of prompting, and limited difficulty level of the generated problems.
To overcome these limitations, we propose \method, a simple yet effective pipeline designed to scale the creation of difficult problems.
We efficiently identify difficult problems from existing datasets with only a single forward pass using an adaptive thinking model, which can perceive problem difficulty and automatically switch between ``Thinking'' and ``NoThinking'' modes.
We then train a specialized difficult problem generator (\gen) on this filtered difficult data, which can produce new difficult problems in large scale, eliminating the need for complex, per-instance prompting and its associated high API costs.
Fine-tuning Qwen2.5-Math-7B-Instruct on the ScaleDiff-Math dataset yields a substantial performance increase of 11.3\% compared to the original dataset and achieves a 65.9\% average accuracy on AIME'24, AIME'25, HMMT-Feb'25, BRUMO'25, and MATH500, outperforming recent strong LRMs like OpenThinker3.
Notably, this performance is achieved using the cost-efficient Qwen3-8B model as a teacher, demonstrating that our pipeline can effectively transfer advanced reasoning capabilities without relying on larger, more expensive teacher models.
Furthermore, we observe a clear scaling phenomenon in model performance on difficult benchmarks as the quantity of difficult problems increases. 
We open-source the \method-Math dataset, the fine-tuned \method model, and the implementation code to facilitate further research and ensure reproducibility.
}

\date{\today}
\correspondence{Conghui He (\email{heconghui@pjlab.org.cn)}, Rui Yan (\email{ruiyan@ruc.edu.cn}), and Lijun Wu (\email{wulijun@pjlab.org.cn})}

\metadata[Model \& Data]{\href{https://huggingface.co/collections/QizhiPei/scalediff-68a71cc18839c1cc1471187e}{ScaleDiff HuggingFace Collection}}
\metadata[Code]{\href{https://github.com/QizhiPei/ScaleDiff}{ScaleDiff Github Repository}}

\begin{document}

\maketitle
\footnotetext[1]{Corresponding Authors.}
\footnotetext[2]{Equal Contribution.}
\footnotetext[3]{Work during internship at Shanghai Artificial Intelligence Laboratory.}

\begin{figure}
    \centering
    \includegraphics[width=0.9\linewidth]{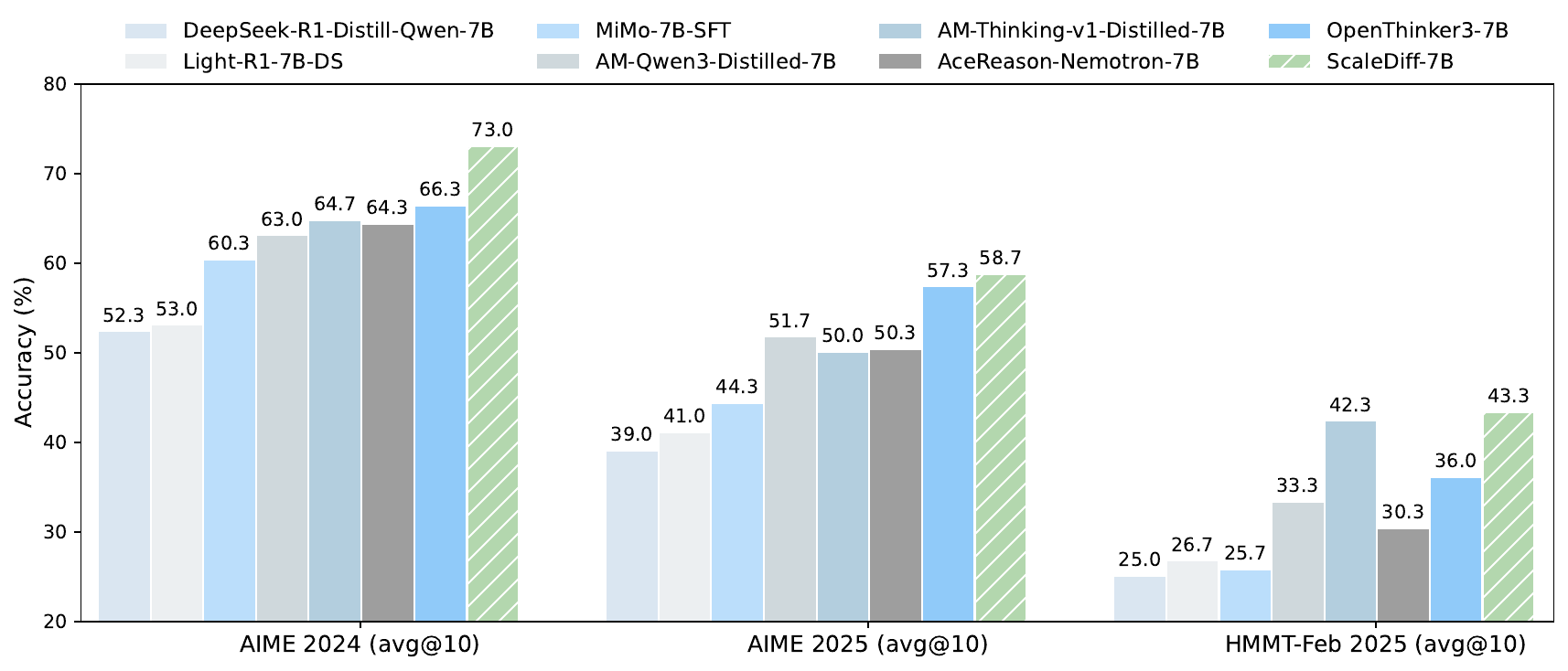}
    \caption{Performance of \method on AIME'24, AIME'25, and HMMT-Feb'25, compared to other SFT-based LRM baselines.}
    \label{fig:placeholder}
\end{figure}

\section{Introduction}

Recent advancements in Large Reasoning Models (LRMs) such as OpenAI-o1~\citep{openai_o1} and DeepSeek-R1~\citep{deepseek_r1} have demonstrated remarkable progress in tackling complex reasoning problems. 
These models exhibit the ability to perform trial-and-error, self-reflection, and iterative refinement within long Chains of Thought (CoT), leading to enhanced problem-solving capabilities.
To replicate this success, various efforts have been made, employing techniques like Supervised Fine-Tuning (SFT) on distilled data~\citep{tian2025not,moshkov2025aimo,openthoughts,guha2025openthoughts3}, Reinforcement Learning (RL) with verifiable rewards~\citep{yu2025dapo,zeng2025simplerl,deepscaler2025,he2025deepmath}, or more complex training pipelines based on SFT and RL~\citep{openr1,wen2025lightr1curriculumsftdpo,acereason_1,acereason_11,rzero}.
A common strategy among these approaches is to identify challenging mathematical problems from existing datasets for training~\citep{moshkov2025aimo,wen2025lightr1curriculumsftdpo,acereason_11}.
The rationale behind this is that difficult problems typically necessitate intricate reasoning processes, thereby stimulating more sophisticated model behaviors, whereas simpler problems often yield limited benefits.
However, creating such difficult mathematical problems—particularly those at the competition or olympiad level—is often costly because they are primarily handcrafted by human experts~\citep{aimo_validation_aime,aime2025,he2024olympiadbench}.
Recent research has explored the automated synthesis of mathematical data by prompting proprietary models as well as large-scale open-source counterparts, either from seed data~\citep{wizardmath,metamath,cot_self_inst,toshniwalopenmathinstruct-2} or from inherent mathematical concepts~\citep{kpmath,mathscale,promptcot,sandmath,zhan2025mathsmith}.
However, scaling these approaches remains challenging due to their substantial computational costs, complex prompting design, and relatively limited difficulty of the generated problems.

To further investigate the impact of difficult problems on enhancing complex reasoning abilities of LRMs, we propose \method, a simple yet effective pipeline that scales the creation of difficult problems to improve models' complex reasoning capabilities.
We begin by leveraging an existing adaptive thinking model~\citep{zhang2025adaptthink}, which can automatically switch between the ``Thinking'' and ``NoThinking'' modes depending on the difficulty of a given problem, thereby serving as a difficult problem identifier to detect difficult problems within existing datasets.
This identification process requires only a single forward pass, making it more efficient than commonly used approaches such as \textit{fail rate} and \textit{LLM-as-a-judge}.
Subsequently, to enable the generation of an arbitrary number of difficult problems, we train a problem generator (denoted as \gen) on these identified difficult problems. 
We then utilize \gen to generate large-scale new difficult problems, eliminating the need for complicated prompting design, per-instance shot selection, and the substantial computational costs required by traditional methods.
For each generated problem, we distill its long CoT solution using Qwen3-8B~\citep{yang2025qwen3} in ``Thinking'' mode.
This comparatively small model provides solutions in a cost-efficient manner and offers a favorable alternative to widely used larger models such as DeepSeek-R1 or QwQ-32B.
We also apply both rule and model filtration to these solutions.
The final \method-Math dataset is composed of these difficult problem-solution pairs and the original dataset. 

Further SFT of Qwen2.5-Math-7B-Instruct on the \method-Math dataset demonstrates promising performance.
Our \method consistently outperforms recent strong LRMs such as OpenThinker3~\citep{guha2025openthoughts3} and AceReason-Nemotron~\citep{acereason_1} across AIME'24, AIME'25, HMMT-Feb'25, BRUMO'25, and MATH500 on average.
It also improves upon AM-Qwen3-Distilled-7B by enhancing both the difficulty and scale of the training data, resulting in a relative performance gain of 11.3\%.
These results highlighting the effectiveness of our approach in strengthing models' complex reasoning abilities.
Moreover, by varying the size of the augmenting dataset, we observe a clear scaling phenomenon in model performance on AIME'24 and AIME'25, with accuracy improving as the number of difficult problems increased. This scaling behavior further highlights the potential of our method to drive continued gains as larger and more challenging datasets become available.

\section{Method}
\begin{figure}
    \centering
    \includegraphics[width=0.9\linewidth]{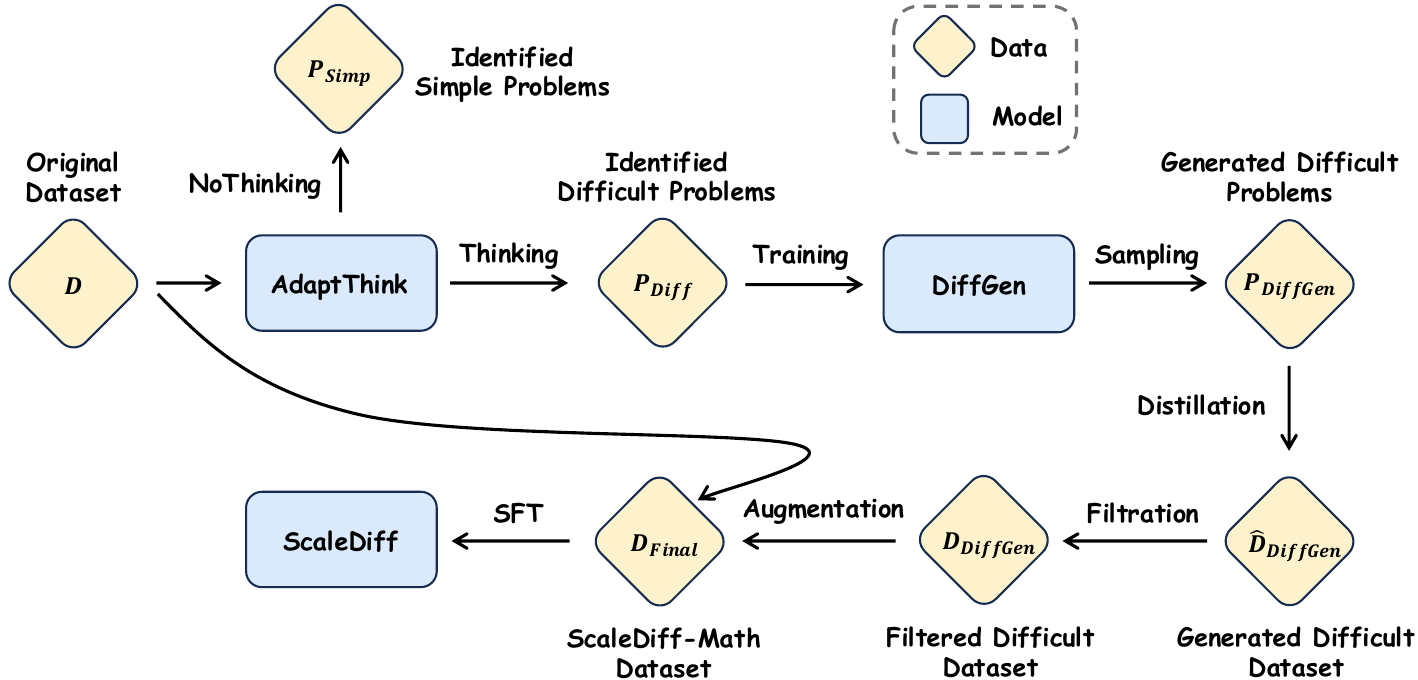}
    \caption{Overview of \method pipeline.}
    \label{fig:overview}
\end{figure}

In this section, we first introduce our identification of difficult problems in Section~\ref{sec:diff_p_identify}.
We then detail our approach for generating a large-scale set of new challenging problems in Section~\ref{sec:diff_p_gen}. 
Finally, in Section~\ref{sec:resp_distill_filt}, we describe our process for distilling and filtering high-quality solutions to these generated problems, which forms the basis of our training dataset.
The overview of our \method pipeline is shown in Figure~\ref{fig:overview}.

\subsection{Difficult Problem Identification}
\label{sec:diff_p_identify}
Assessing the difficulty of mathematical problems primarily relies on two existing methods: \textit{fail rate}~\citep{dartmath} and \textit{LLM-as-a-judge}~\citep{omni_math}.
Specifically, for \textit{fail rate}, a proxy mathematical model is used to solve a given problem multiple times, and the proportion of incorrect responses determines its fail rate. 
For \textit{LLM-as-a-judge}, a more powerful LLM is prompted with the mathematical problem, its reference solution (if available), and predefined criteria for difficulty assessment.
However, both methods have their limitations: the \textit{fail rate} is computationally inefficient as it necessitates multiple solution attempts by the proxy mathematical model;
\textit{LLM-as-a-judge} is highly sensitive to the specific rules and criteria predefined within the input prompt.

Different from existing methods, we seek to leverage \textit{AdaptThink}\footnote{\url{https://huggingface.co/THU-KEG/AdaptThink-7B-delta0.05}}~\citep{zhang2025adaptthink} as our difficult problem identifier. \textit{AdaptThink} algorithm is designed to teach models to adaptively choose between a time-consuming ``Thinking'' process for complex problems and a direct ``NoThinking'' response for simpler ones through RL. This adaptive mechanism inherently reflects the model's perceived difficulty of a problem.
The primary objective of \textit{AdaptThink} is a constrained optimization:
\begin{equation}
    \text{max } \mathbb{E}_{(x,\cdot)\sim\mathcal{D},y\sim\pi_{\theta}(\cdot|x)}\mathbb{I}(y_{1}=\verb|</think>|), \text{ s.t. } \mathbb{E}_{(x,\cdot)\sim\mathcal{D},y\sim\pi_{\theta}(\cdot|x)}R(x,y)\ge \mathbb{E}_{(x,\cdot)\sim\mathcal{D},y^{\prime}\sim\pi_{\theta_{ref}}(\cdot|x)}R(x,y^{\prime}),
\end{equation}
where $\mathcal{D}$ denotes the problem-solution dataset, and we let $\mathcal{P}=\{x \mid (x,y)\in\mathcal{D}\}$ denote its problem set. $\mathbb{I}(y_{1}=\verb|</think>|)$ is an indicator function for the first generated token being \verb|</think>|. $R(x,y)$ is the reward function representing the accuracy of the model's response for problem $x$ (returning 1 for a correct solution and 0 for an incorrect one).
This objective aims to maximize the probability of generating ``NoThinking'' responses, subject to a constraint: the current model's expected accuracy---the reward for a correct solution $R(x,y)$---must be maintained at or above that of a fixed reference model.
This design compels \textit{AdaptThink} to opt for the efficient ``NoThinking'' mode only when it does not compromise accuracy.
Conversely, for problems where ``NoThinking'' would lead to a significant performance drop, \textit{AdaptThink} is driven to engage its ``Thinking'' mode to satisfy the accuracy constraint.

This adaptive behavior effectively transforms \textit{AdaptThink} into a binary classifier for problem difficulty. We define a problem $x$ as ``simple'' if \textit{AdaptThink} produces a ``NoThinking'' response, and ``difficult'' otherwise.
Formally, our problem identification criteria are based on the first generated token ($y_1$) of \textit{AdaptThink}'s response:
\begin{equation}
\text{Difficulty}(x) = 
\begin{cases}
    \text{Simple} & \text{if } y_1 = \verb|</think>|\\
    \text{Difficult} & \text{if } y_1 \neq \verb|</think>|\\
\end{cases}
\end{equation}
Notably, determining whether a problem is simple or difficult requires only the model's output of a first token (one forward pass), making the identification process highly efficient than \textit{fail rate} and \textit{LLM-as-a-judge}. By applying \textit{AdaptThink} as such a identifier, we efficiently identify and extract challenging problem-solution pairs from existing datasets, forming a curated subset denoted as $\mathcal{D}_{\text{Diff}}$. 

\subsubsection{Effectiveness of Difficulty}
\label{sec:simp_vs_diff}

To evaluate the validity of using \textit{AdaptThink} as a difficult problem identifier and the effectiveness of difficult problems as training data, we conduct SFT on Qwen2.5-Math-7B-Instruct~\citep{yang2024qwen2math} with the full dataset $\mathcal{D}$, as well as its two subsets: the identified difficult subset $\mathcal{D}_{\text{Diff}}$ and simple subset $\mathcal{D}_{\text{Simp}}$. 
To further control for data size, we downsample $\mathcal{D}_{\text{Simp}}$ to match the size of $\mathcal{D}_{\text{Diff}}$ and additionally construct a random subset $\mathcal{D}_{\text{Rand}}$ by sampling 192K problems from $\mathcal{D}$.
We use the AM-Qwen3-Distilled dataset as our initial $\mathcal{D}$, with more experimental details provided in Section~\ref{sec:exp_setup}.

\begin{table}[h!]
\renewcommand{\arraystretch}{1.1}
\centering
\resizebox{\linewidth}{!}{
\begin{tabular}{lcccccc|c}
\shline
\multirow{2}{*}{\textbf{Model}} & \multirow{2}{*}{\textbf{Size}} & \textbf{AIME'24} & \textbf{AIME'25} & \textbf{HMMT-Feb'25} & \textbf{BRUMO'25} & \textbf{MATH500} & \multirow{2}{*}{\textbf{AVG}} \\
& & \textcolor{gray}{avg@10} & \textcolor{gray}{avg@10} & \textcolor{gray}{avg@10} & \textcolor{gray}{avg@10} & \textcolor{gray}{avg@3}  \\ \shline
$\mathcal{D}$ & 558K & 63.0{\scriptsize$\pm$3.5} & 51.7{\scriptsize$\pm$5.6} & 33.3{\scriptsize$\pm$5.8} & 60.7{\scriptsize$\pm$7.7} & 94.6{\scriptsize$\pm$0.4} & 59.2 \\ 
$\mathcal{D}_{Simp}$ & 366K & 43.7{\scriptsize$\pm$5.3} & 38.7{\scriptsize$\pm$4.5} & 27.0{\scriptsize$\pm$5.3} & 53.7{\scriptsize$\pm$6.0} & 91.3{\scriptsize$\pm$0.4} & 48.9 \\ 
$\mathcal{D}_{Simp}$ & 192K & 40.7{\scriptsize$\pm$4.9} & 33.7{\scriptsize$\pm$2.8} & 24.0{\scriptsize$\pm$3.6} & 48.3{\scriptsize$\pm$7.8} & 90.4{\scriptsize$\pm$0.7} & 45.1 \\ 
$\mathcal{D}_{Rand}$ & 192K & 54.3{\scriptsize$\pm$5.0} & 42.0{\scriptsize$\pm$5.2} & 31.3{\scriptsize$\pm$4.8} & 57.0{\scriptsize$\pm$4.6} & 93.2{\scriptsize$\pm$0.4} & 53.3 \\ 
$\mathcal{D}_{Diff}$ & 192K & 62.3{\scriptsize$\pm$5.0} & 44.3{\scriptsize$\pm$7.6} & 36.0{\scriptsize$\pm$5.7} & 59.0{\scriptsize$\pm$6.3} & 93.9{\scriptsize$\pm$1.2} & 56.6 \\ 
\shline
\end{tabular}
}
\caption{Effect of problem difficulty on SFT performance across three mathematical benchmarks.}
\label{tab:simp_diff}
\end{table}

The corresponding data size and results on three mathematical benchmarks are presented in Table~\ref{tab:simp_diff}. It can be observed that training on the difficult subset $\mathcal{D}_{\text{Diff}}$ (192K) outperforms training on the simple subset $\mathcal{D}_{\text{Simp}}$ (56.6 vs. 45.1 on average) and randomly sampled subset $\mathcal{D}_{\text{Rand}}$ (56.6 vs. 53.3) of the same size , highlighting that difficult problems provide more effective training signals for enhancing reasoning ability. 
Even when comparing $\mathcal{D}_{\text{Diff}}$ (192K) against the much larger $\mathcal{D}_{\text{Simp}}$ (366K), the difficult subset still yields a clear advantage (56.6 vs. 48.9). 
Moreover, while training on the full dataset $\mathcal{D}$ (558K) achieves the strongest results overall (59.2), this improvement stems primarily from its larger scale. Notably, the performance gap between $\mathcal{D}_{\text{Diff}}$ (192K) and the full dataset is only 2.6 points (56.6 vs. 59.2), despite the latter being nearly three times larger. In contrast, the gap between $\mathcal{D}_{\text{Simp}}$ (192K) and the full dataset is approximately 14 points (45.1 vs. 59.2), underscoring that simple problems contribute far less effectively to improving model performance compared to difficult ones.
These findings confirm that \textit{AdaptThink} serves as an effective identifier for identifying high-value difficult problems, and that such problems are significantly more beneficial than simple ones in improving model performance.

\subsection{Difficult Problem Generator}
\label{sec:diff_p_gen}
Building upon the identified difficult problem set $\mathcal{P}_{\text{Diff}}$ from $\mathcal{D}_{\text{Diff}}$ in Section~\ref{sec:diff_p_identify}, we train a dedicated difficult problem generator, denoted as \gen, following a similar methodology to ScaleQuest~\citep{scale_quest}.
The rationale for training exclusively on difficult problem sets derives from Section~\ref{sec:simp_vs_diff}, which demonstrates that difficult problems are more effective than simple ones in improving model performance.

For each problem $x=(x_1, x_2, \dots, x_n)$ in $\mathcal{P}_{\text{Diff}}$, where $x_i$ denotes the $i_{th}$ token of the problem, we construct the input by prepending a standard instruction prefix $\mathcal{I} = \texttt{<|im\_start|>user}\verb|\n|$.
Distinct from conventional instruction tuning---where the loss is often computed over solution tokens---our training loss is only applied to the problem tokens $x_i$ without solution as input.
The training objective for \gen is the standard cross-entropy loss for language modeling:
\begin{equation}
    \mathcal{L}_{\text{CE}}(\theta) = -\frac{1}{n} \sum_{i=1}^{n} \log P(x_i | \mathcal{I}, x_1, \dots, x_{i-1}).
\end{equation}
This design encourages \gen to capture the distributional patterns inherent in challenging mathematical problems. Importantly, the goal is not to optimize for problem solving, but rather to enable the model to generate new problems of comparable complexity.

Given the instruction prefix $\mathcal{I}$, the trained generator \gen can produce a large number of new difficult problems by adjusting decoding parameters such as temperature and top-p. 
The resulting collection of generated problems constitutes our problem set, denoted as $\mathcal{P}_{\text{DiffGen}}$.

\subsection{Solution Distillation and Filtration}
\label{sec:resp_distill_filt}
After generating the problem set $\mathcal{P}_{\text{DiffGen}}$, we re-evaluate their difficulty using the methodology introduced in Section~\ref{sec:diff_p_identify}. 
The validation shows that about 88\% of the generated problems are classified as difficult, suggesting that \gen effectively captures the distributional characteristics of challenging problems.
Since assessing the mathematical correctness and solvability of generated problems remains a highly non-trivial task, we leave this aspect as future work and focus instead on ensuring the quality of distilled solutions.

For each problem in $\mathcal{P}_{\text{DiffGen}}$, we utilize a strong teacher model to distill its corresponding solution, resulting in $\hat{\mathcal{D}}_{\text{DiffGen}}$.
Upon obtaining these solutions, we perform a two-stage filtration process: rule and model filtration. 
The initial rule filtering stage removes solutions with common undesirable traits. 
This includes cases with extensive repetition or overly verbose reasoning that prevents the final answer from being clearly encapsulated within \verb|\boxed{}|.
The model filtering further refines the dataset by discarding problems that our base model can already solve reliably. Specifically, if the base model's predicted answer matches the teacher-provided answer, the problem is considered redundant and removed from the dataset.
In total, we filter out approximately 43\% of the initial samples and obtain the final problem-solution dataset $\mathcal{D}_{\text{DiffGen}}$.

\section{Experiments}
\label{sec:exp}
\subsection{Experimental Setup}
\label{sec:exp_setup}
\textbf{\noindent Initial Dataset $\mathcal{D}$.}
We utilize the mathematical domain subset of the AM-Qwen3-Distilled dataset\footnote{\url{https://huggingface.co/datasets/a-m-team/AM-Qwen3-Distilled}} as our initial dataset $\mathcal{D}$, which is  well-regarded for its high quality and has demonstrated effectiveness in training mathematical reasoning models.
Its problem set $\mathcal{P}$ is a compilation of several prominent sub-datasets, including DeepMath-103K~\citep{he2025deepmath}, OpenR1-Math-220K~\citep{openr1}, OpenMathReasoning~\citep{moshkov2025aimo}, and NuminaMath~\citep{li2024numinamath}, among others.
Subsequently, $\mathcal{P}$ undergoes rigorous deduplication, rule filtering, and decontamination concerning downstream tasks.
The original solutions in $\mathcal{D}$ are distilled from Qwen3-235B-A22B~\citep{yang2025qwen3}.
This distillation process is iteratively repeated until a correct solution is obtained. Further filtering is also conducted based on metrics such as perplexity (PPL) and Ngram scores~\citep{tian2025not}.
This multi-stage processing results in a final curated dataset comprising 558K data instances.

\textbf{\noindent Training of \gen.}
Following the identification process described in Section~\ref{sec:diff_p_identify}, 192K problems from $\mathcal{P}$ are classified as difficult, denoted as $\mathcal{P}_{\text{Diff}}$. 
We train \gen based on the Qwen3-8B-Base~\citep{yang2025qwen3} model. 
The training configuration consists of a batch size of 128, a maximum sequence length of 1024 tokens, a learning rate of 5e-5, and a total of 1 epoch.
We employ 10\% warmup steps with a linear decay learning rate schedule.
We use LLaMA-Factory~\citep{zheng2024llamafactory} as our training framework.

\begin{table}[t]
\renewcommand{\arraystretch}{1.1}
\centering
\resizebox{\linewidth}{!}{
\begin{tabular}{lccccc|c}
\shline
\multirow{2}{*}{\textbf{Model}} & \textbf{AIME'24} & \textbf{AIME'25} & \textbf{HMMT-Feb'25} & \textbf{BRUMO'25} & \textbf{MATH500} & \multirow{2}{*}{\textbf{AVG}} \\
& \textcolor{gray}{avg@10} & \textcolor{gray}{avg@10} & \textcolor{gray}{avg@10} & \textcolor{gray}{avg@10} & \textcolor{gray}{avg@3}  \\ \shline
Qwen2.5-7B-Instruct~\citep{yang2024qwen2} & 11.3{\scriptsize$\pm$5.4} & 11.0{\scriptsize$\pm$5.2} & 2.7{\scriptsize$\pm$2.0} & 22.3{\scriptsize$\pm$3.7} & 77.5{\scriptsize$\pm$1.0} & 22.6 \\
Qwen2.5-Math-7B-Instruct~\citep{yang2024qwen2math} & 11.3{\scriptsize$\pm$2.7} & 11.3{\scriptsize$\pm$3.1} & 2.0{\scriptsize$\pm$1.6} & 18.0{\scriptsize$\pm$6.0} & 82.7{\scriptsize$\pm$0.2} & 22.8 \\
DeepSeek-R1-Distill-Qwen-7B~\citep{deepseekai2025} & 53.0{\scriptsize$\pm$5.3} & 41.7{\scriptsize$\pm$6.5} & 25.0{\scriptsize$\pm$3.7} & 54.7{\scriptsize$\pm$7.2} & 93.7{\scriptsize$\pm$0.4} & 51.6 \\
Qwen3-8B~\citep{yang2025qwen3} & 75.0{\scriptsize$\pm$4.5} & 64.7{\scriptsize$\pm$6.4} & 44.0{\scriptsize$\pm$4.4} & 68.0{\scriptsize$\pm$2.7} & 96.8{\scriptsize$\pm$0.3} & 68.9 \\

\shline
\rowcolor{blue!15}\multicolumn{7}{c}{\textbf{RL}} \\
\shline
LIMR-7B~\citep{li2025limr} & 33.3{\scriptsize$\pm$4.5} & 7.3{\scriptsize$\pm$3.3} & 0.7{\scriptsize$\pm$1.3} & 20.3{\scriptsize$\pm$4.3} & 77.4{\scriptsize$\pm$0.6} & 24.4 \\
Oat-Zero-7B~\citep{liu2025oatzero} & 30.0{\scriptsize$\pm$4.5} & 11.0{\scriptsize$\pm$4.0} & 4.0{\scriptsize$\pm$2.9} & 22.0{\scriptsize$\pm$3.7} & 79.4{\scriptsize$\pm$0.3} & 26.2 \\
Open-Reasoner-Zero-7B~\citep{open_reasoner_zero} & 17.0{\scriptsize$\pm$3.5} & 17.0{\scriptsize$\pm$3.1} & 3.0{\scriptsize$\pm$2.3} & 29.3{\scriptsize$\pm$2.9} & 82.4{\scriptsize$\pm$1.3} & 27.6 \\
AReaL-boba-RL-7B~\cite{areal} & 58.0{\scriptsize$\pm$4.8} & 43.0{\scriptsize$\pm$4.8} & 25.3{\scriptsize$\pm$4.0} & 56.3{\scriptsize$\pm$5.8} & 93.2{\scriptsize$\pm$0.6} & 53.1 \\
Skywork-OR1-Math-7B~\cite{he2025skywork} & 59.7{\scriptsize$\pm$3.8} & 49.7{\scriptsize$\pm$5.0} & 30.3{\scriptsize$\pm$4.8} & 61.7{\scriptsize$\pm$4.5} & 95.3{\scriptsize$\pm$0.1} & 57.7 \\
Skywork-OR1-7B~\cite{he2025skywork} & 61.5{\scriptsize$\pm$4.2} & 50.3{\scriptsize$\pm$5.5} & 28.0{\scriptsize$\pm$5.0} & 63.7{\scriptsize$\pm$6.0} & 95.9{\scriptsize$\pm$0.2} & 58.3 \\
MiMo-7B-RL~\cite{mimo} & 68.3{\scriptsize$\pm$4.3} & 59.0{\scriptsize$\pm$5.0} & 38.3{\scriptsize$\pm$4.8} & 64.3{\scriptsize$\pm$2.6} & 95.6{\scriptsize$\pm$0.4} & 64.1 \\

\shline
\rowcolor{blue!15}\multicolumn{7}{c}{\textbf{SFT}} \\
\shline
OpenThinker-7B~\citep{openthoughts} & 28.0{\scriptsize$\pm$4.3} & 25.7{\scriptsize$\pm$4.7} & 18.0{\scriptsize$\pm$5.8} & 36.7{\scriptsize$\pm$4.7} & 87.9{\scriptsize$\pm$0.4} & 37.0 \\
OpenR1-Qwen-7B~\citep{openr1} & 50.7{\scriptsize$\pm$5.1} & 36.3{\scriptsize$\pm$3.5} & 25.7{\scriptsize$\pm$3.0} & 55.7{\scriptsize$\pm$6.2} & 93.4{\scriptsize$\pm$0.7} & 49.7 \\
OpenThinker2-7B~\citep{openthoughts} & 54.7{\scriptsize$\pm$7.6} & 38.0{\scriptsize$\pm$5.6} & 23.0{\scriptsize$\pm$4.1} & 54.7{\scriptsize$\pm$4.3} & 93.9{\scriptsize$\pm$0.4} & 50.4 \\
Light-R1-7B-DS~\citep{wen2025lightr1curriculumsftdpo} & 55.3{\scriptsize$\pm$5.4} & 41.3{\scriptsize$\pm$2.7} & 26.7{\scriptsize$\pm$3.7} & 56.0{\scriptsize$\pm$4.9} & 94.0{\scriptsize$\pm$0.3} & 52.4 \\
MiMo-7B-SFT~\citep{mimo} & 60.3{\scriptsize$\pm$6.0} & 44.3{\scriptsize$\pm$6.7} & 25.7{\scriptsize$\pm$4.5} & 50.7{\scriptsize$\pm$8.1} & 93.6{\scriptsize$\pm$0.2} & 53.2 \\
AceReason-Nemotron-7B~\citep{acereason_1} & 64.3{\scriptsize$\pm$2.6} & 50.3{\scriptsize$\pm$2.8} & 30.3{\scriptsize$\pm$3.5} & 63.7{\scriptsize$\pm$6.0} & 96.1{\scriptsize$\pm$0.4} & 59.2 \\
\rowcolor{gray!30}AM-Qwen3-Distilled-7B$^*$~\citep{tian2025not} & 63.0{\scriptsize$\pm$3.5} & 51.7{\scriptsize$\pm$5.6} & 33.3{\scriptsize$\pm$5.8} & 60.7{\scriptsize$\pm$7.7} & 94.6{\scriptsize$\pm$0.4} & 59.2 \\
AM-Thinking-v1-Distilled-7B$^*$~\citep{tian2025not}  & 62.0{\scriptsize$\pm$5.8} & 50.0{\scriptsize$\pm$3.3} & 42.3{\scriptsize$\pm$4.0} & 62.7{\scriptsize$\pm$3.9} & 94.9{\scriptsize$\pm$0.7} & 60.3 \\
OpenThinker3-7B~\citep{guha2025openthoughts3} & 66.3{\scriptsize$\pm$4.3} & 57.3{\scriptsize$\pm$5.5} & 36.0{\scriptsize$\pm$3.9} & 67.7{\scriptsize$\pm$3.0} & 95.8{\scriptsize$\pm$0.4} & 63.4 \\
OpenMath-Nemotron-7B~\citep{moshkov2025aimo} & 73.7{\scriptsize$\pm$4.1} & 60.7{\scriptsize$\pm$4.7} & 43.0{\scriptsize$\pm$5.5} & 68.0{\scriptsize$\pm$6.2} & 95.2{\scriptsize$\pm$0.3} & 66.9 \\
\shline
\method-7B & 73.0{\scriptsize$\pm$5.0} & 58.7{\scriptsize$\pm$8.2} & 43.3{\scriptsize$\pm$4.2} & 66.7{\scriptsize$\pm$2.7} & 95.2{\scriptsize$\pm$0.3} & 65.9 \\
\shline
\end{tabular}
}
\caption{Pass@1 accuracy (mean $\pm$ std) comparison of different LRMs on AIME'24, AIME'25, HMMT-Feb'25, BRUMO'25, and MATH500 benchmarks with multiple runs.
The baseline results are sorted by the average performance.
$^*$ denotes results from our evaluation of the Qwen2.5-Math-7B-Instruct model trained by us on the corresponding dataset.
The rows highlighted in gray correspond to the source data $\mathcal{D}$ used for the \method augmentation.
}
\label{tab:accuracy_results}
\end{table}
\textbf{\noindent Construction of $\mathcal{D}_{\text{DiffGen}}$.}
Upon completion of training, we use \gen to generate the $\mathcal{P}_{\text{DiffGen}}$. Generation parameters are set to a temperature of 1.0, a top-p value of 0.95, and a top-k value of 20.
We utilize the Qwen3-8B model~\citep{yang2025qwen3} as teacher model to generate long CoT solutions for the problems within $\mathcal{P}_{\text{DiffGen}}$ in ``Thinking'' mode, resulting in $\hat{\mathcal{D}}_{\text{DiffGen}}$.
These generated solutions then undergo the filtration process detailed in Section~\ref{sec:resp_distill_filt}. 
For the model filtering stage, we specifically employ the Qwen2.5-Math-7B-Instruct~\citep{qwen25_math} model, given that this model also serves as the base model for our \method. 
This comprehensive process yields our generated dataset, denoted as $\mathcal{D}_{\text{DiffGen}}$, comprising 1.15M problem-solution pairs.
By augmenting $\mathcal{D}$ with $\mathcal{D}_{\text{DiffGen}}$, we get the final $\mathcal{D}_{Final}$ (\method-Math) dataset, comprising 1.7M problem-solution pairs.

\textbf{\noindent Training of \method.}
As described above, in SFT, \method model is initialized from Qwen2.5-Math-7B-Instruct~\citep{qwen25_math} model and trained on \method-Math dataset. 
The batch size is set to 32, the maximum sequence length is 32,768 tokens, the training epoch is set to 3, with other training settings consistent with those employed for training \gen.
Due to the native context length limitation of the Qwen2.5-Math-7B-Instruct model to 4,096 tokens, we modify the rope\_theta parameter from 10K to 300K to enable support for a maximum context length of 32,768 tokens, following the practice of OpenR1~\citep{openr1}.
The data template used for fine-tuning follows the default format of Qwen series.

\textbf{\noindent Evaluation.}
To ensure robust and reproducible results, our evaluation adheres to the standardized framework and best practices outlined in~\citep{hochlehnert2025sober}. 
We assess the performance of our \method model against relevant baselines on a comprehensive set of widely recognized mathematical reasoning benchmarks:  
AIME'24~\citep{aimo_validation_aime}, AIME'25~\citep{aime2025}, HMMT Feb'25~\citep{hummt}, BRUMO~\citep{brumo}, and MATH500~\citep{huggingfaceh4-math500}.
Performance is primarily measured using the standard Pass@1 metric.
To account for potential variability, especially on smaller benchmarks, all evaluation results are averaged over multiple random seeds. Specifically, we use 10 random seeds for AIME'24, AIME'25, HMMT-Feb'25, BRUMO'25, and 3 random seeds for MATH500.
The maximum number of new tokens, temperature, and top-p are set to 32,768, 0.6, and 0.95, respectively.
All evaluations are conducted using the LightEval framework~\citep{lighteval} with a vLLM backend~\citep{kwon2023vllm} and default chat template.

All training and evaluation detailed above were conducted on NVIDIA A800 GPUs.

\textbf{\noindent Baselines.}
We mainly compare \method with Qwen2.5-7B model series, including Qwen2.5-7B-Instruct~\citep{yang2024qwen2}, Qwen2.5-Math-7B-Instruct~\citep{yang2024qwen2math}, DeepSeek-R1-Distill-Qwen-7B~\citep{deepseek_r1}, as well as LRMs that have undergone further SFT or RL based on Qwen2.5-7B model series.

\subsection{Main Results}

Our \method demonstrates strong performance on both relatively simple benchmark MATH500 and more challenging benchmarks including AIME, HMMT-Feb'25, and BRUMO, achieving average accuracies that surpass many RL- or SFT-based strong LRMs, such as MiMo-7B-RL~\citep{mimo}, Light-R1-7B-DS~\citep{wen2025lightr1curriculumsftdpo}, AceReason-Nemotron-7B~\citep{acereason_1}, and the recent OpenThinker3-7B~\citep{guha2025openthoughts3}.
Unlike most of these baseline methods, which rely on rejection sampling during solution distillation—sampling multiple candidate solutions and retaining only those matching the ground-truth answer—our approach samples a single response per problem. This eliminates the need for repeated sampling until the correct solution is found, resulting in significantly lower data generation cost. 
Although the training data may contain incorrect answers, the diverse reasoning traces they provide can still contribute to enhancing the model's reasoning ability. This observation is consistent with prior findings reported in~\citep{toshniwalopenmathinstruct-2,su2025klear}.

\textbf{\noindent Comparison with AM-Qwen3-Distilled-7B.}
\method achieves substantial improvements (11.3\%) over AM-Qwen3-Distilled-7B~\citep{tian2025not},
as \method-7B can be viewed as a ``hiking'' version of AM-Qwen3-Distilled-7B. Here hiking refers to increasing both the overall difficulty and volume of the dataset through the \method pipeline.
We believe that such difficulty hiking is generally applicable when the original dataset maintains a balanced difficulty distribution.

\textbf{\noindent Comparison with teacher model Qwen3-8B.}
Qwen3-8B is the teacher model for \method-7B. From the results in Table \ref{tab:accuracy_results}, \method-7B achieves 65.9\% average accuracy, which closely approaches Qwen3-8B's 68.9\%. The gap between the two models is thus relatively small overall, indicating that the distillation and difficulty-hiking pipeline successfully transfers much of the teacher's reasoning ability into the student model.

\subsection{Ablation Study}
We further conduct an ablation study to investigate the contributions of different components in the \method pipeline. Specifically, we focus on two key modules: (1) difficult problem identification and (2) response filtration.

To verify the effect of difficult problem identification, we remove both the identification and the subsequent filtration steps, and instead train the question generator directly on the original problem set $\mathcal{P}$. We then generate new problems from this generator, distill responses from the same teacher model, and fine-tune the same target model.
To assess the effect of response filtration, we keep the difficult problem identification step but remove the filtration.
For fair comparison, the total fine-tuning data size is fixed to $192$K samples across all experiments. The results are summarized in Table~\ref{tab:ablation}, from which we can observe:
\begin{table}[h!]
\renewcommand{\arraystretch}{1.1}
\centering
\resizebox{\linewidth}{!}{
\begin{tabular}{lcccccc|c}
\shline
\multirow{2}{*}{\textbf{Model}} & \multirow{2}{*}{\textbf{Size}} & \textbf{AIME'24} & \textbf{AIME'25} & \textbf{HMMT-Feb'25} & \textbf{BRUMO'25} & \textbf{MATH500} & \multirow{2}{*}{\textbf{AVG}} \\
& & \textcolor{gray}{avg@10} & \textcolor{gray}{avg@10} & \textcolor{gray}{avg@10} & \textcolor{gray}{avg@10} & \textcolor{gray}{avg@3}  \\ \shline
\method & 192K & 61.0{\scriptsize$\pm$5.2} & 52.0{\scriptsize$\pm$5.0} & 33.0{\scriptsize$\pm$3.5} & 57.7{\scriptsize$\pm$5.0} & 94.7{\scriptsize$\pm$0.1} & 58.4 \\ 
\shline
w/o Filtration & 192K & 59.0{\scriptsize$\pm$7.5} & 46.7{\scriptsize$\pm$7.3} & 29.3{\scriptsize$\pm$7.9} & 56.7{\scriptsize$\pm$4.9} & 93.3{\scriptsize$\pm$0.5} & 55.3 \\ 
w/o Filtration \& Difficult & 192K & 47.7{\scriptsize$\pm$5.8} & 45.0{\scriptsize$\pm$6.2} & 25.0{\scriptsize$\pm$3.4} & 47.0{\scriptsize$\pm$3.5} & 92.5{\scriptsize$\pm$0.9} & 50.4 \\ 
\shline
\end{tabular}
}
\caption{Ablation Study on the effects of difficult problem selection and response filtration.}
\label{tab:ablation}
\end{table}
(1) Removing response filtration degrades performance (58.4 $\rightarrow$ 55.3 on average), showing that both rule and model filtering are important to eliminate noisy, repetitive, or low-value samples. This ensures the fine-tuning dataset remains both high-quality and challenging.
(2) Removing difficult problem identification further causes a notable drop in performance (55.3 $\rightarrow$ 50.4), confirming that pre-filtering challenging problems before generator training yields more effective data for enhancing reasoning capabilities. Without this step, the generated dataset may contain a higher proportion of trivial problems, limiting SFT gains.

\section{Analysis}
In this section, we present a series of analyses to investigate the impact of data scaling (Section~\ref{sec:analy_scale}), the effect of teacher model (Section~\ref{sec:analy_teacher}), and the difficulty of generated problems (Section~\ref{sec:analy_dist}). 
Unless otherwise specified, all experiments are conducted on unfiltered solutions.

\subsection{Impact of Data Scaling}
\label{sec:analy_scale}
\begin{figure}
    \centering
    \includegraphics[width=0.9\linewidth]{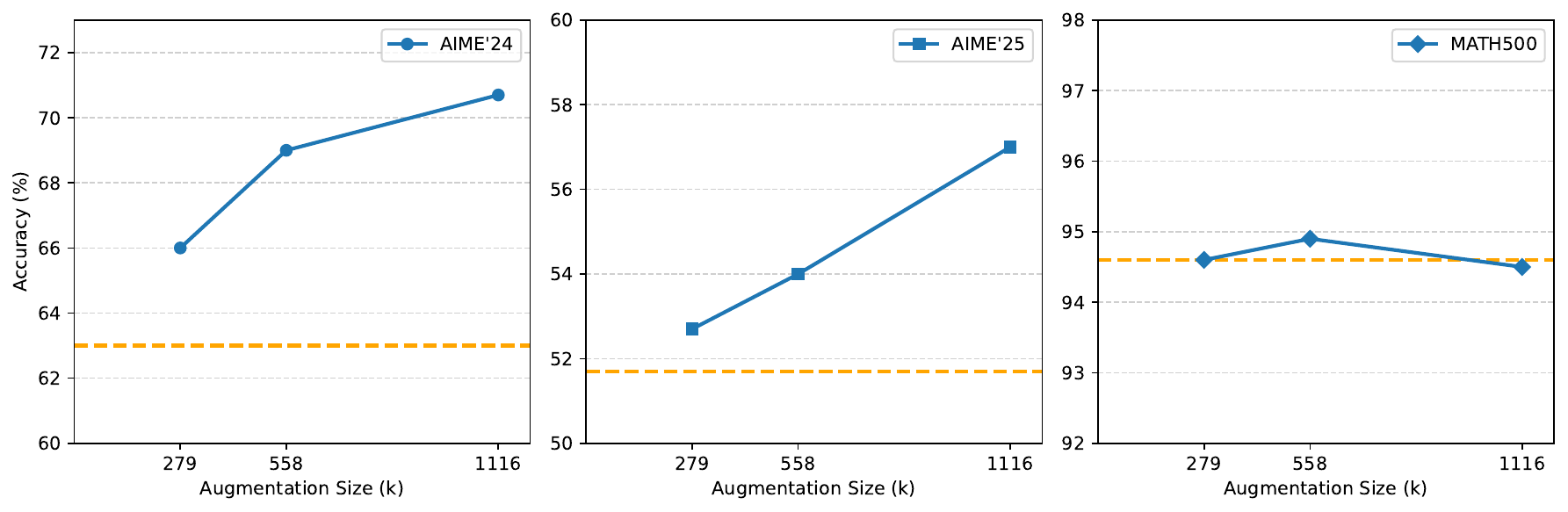}
    \caption{Accuracy scaling with the size of augmented data on AIME'24, AIME'25, and MATH500. The amount of augmented data is 1/2, 1, and 2 times the size of the original dataset.}
    \label{fig:acc_scaling}
\end{figure}
To assess the impact of augmentation scale on downstream performance, we vary the size of the generated dataset and evaluate the model across 3 benchmarks.
Figure~\ref{fig:acc_scaling} illustrates the effect of augmentation dataset size on model performance across three benchmarks: AIME'24, AIME'25, and MATH500. The yellow dashed line denotes the baseline results of AM-Qwen3-Distilled-7B without augmentation.
From the figure, we observe a consistent performance improvement on the more challenging AIME'24 and AIME'25 benchmarks as the augmentation dataset size increases. Notably, even when the augmentation size reaches twice that of the original dataset, performance gains remain unsaturated, indicating the continued benefit of scaling difficult problems for enhancing complex reasoning.
In contrast, for the relatively easier MATH500 benchmark, the augmentation provides no improvements, suggesting that additional difficult data contributes more significantly when the evaluation tasks themselves demand complex reasoning.

\subsection{Effect of Teacher Model}
\label{sec:analy_teacher}

The performance of different teacher models may vary, and consequently, the quality of their distilled responses can influence downstream results. In this section, we investigate the effect of the teacher model.
In our pipeline, the original solutions for $\mathcal{D}_{\text{Diff}}$ are distilled from Qwen3-235B-A22B. 
We further distill three responses for each problem in $\mathcal{P}_{\text{Diff}}$ using Qwen3-8B.
We then compare the results obtained from the two teacher models on this controlled dataset. 
\begin{table}[h!]
\renewcommand{\arraystretch}{1.1}
\centering
\resizebox{\linewidth}{!}{
\begin{tabular}{lcccccc|c}
\shline
\multirow{2}{*}{\textbf{Teacher Model}} & \multirow{2}{*}{\textbf{Size}} & \textbf{AIME'24} & \textbf{AIME'25} & \textbf{HMMT-Feb'25} & \textbf{BRUMO'25} & \textbf{MATH500} & \multirow{2}{*}{\textbf{AVG}} \\
& & \textcolor{gray}{avg@10} & \textcolor{gray}{avg@10} & \textcolor{gray}{avg@10} & \textcolor{gray}{avg@10} & \textcolor{gray}{avg@3}  \\ \shline
Qwen3-235B-A22B & 192K & 62.3{\scriptsize$\pm$5.0} & 44.3{\scriptsize$\pm$7.6} & 36.0{\scriptsize$\pm$5.7} & 59.0{\scriptsize$\pm$6.3} & 93.9{\scriptsize$\pm$1.2} & 56.6 \\
Qwen3-8B & 192K & 57.3{\scriptsize$\pm$5.3} & 50.0{\scriptsize$\pm$6.7} & 26.7{\scriptsize$\pm$6.3} & 56.3{\scriptsize$\pm$6.6} & 93.5{\scriptsize$\pm$1.0} & 55.6 \\
\shline
\end{tabular}
}
\caption{The effect of teacher model for solution distillation.}
\label{tab:teacher_model}
\end{table}
As shown in Table~\ref{tab:teacher_model}, we observe that using Qwen3-235B-A22B as the teacher model yields slightly better performance than Qwen3-8B, though the difference is not substantial. This finding partially aligns with prior observations in~\citep{guha2025openthoughts3,li2025small}, which suggest that stronger-performing models are not necessarily better ``teachers'' because a noticeable gap often exists between large teacher models and smaller student models.
These results corroborate our decision to adopt the smaller Qwen3-8B as a teacher model, demonstrating it to be a more cost-efficient choice.

\subsection{Difficulty of Generated Problems}
\label{sec:analy_dist}

\begin{figure}
    \centering
    \includegraphics[width=0.7\linewidth]{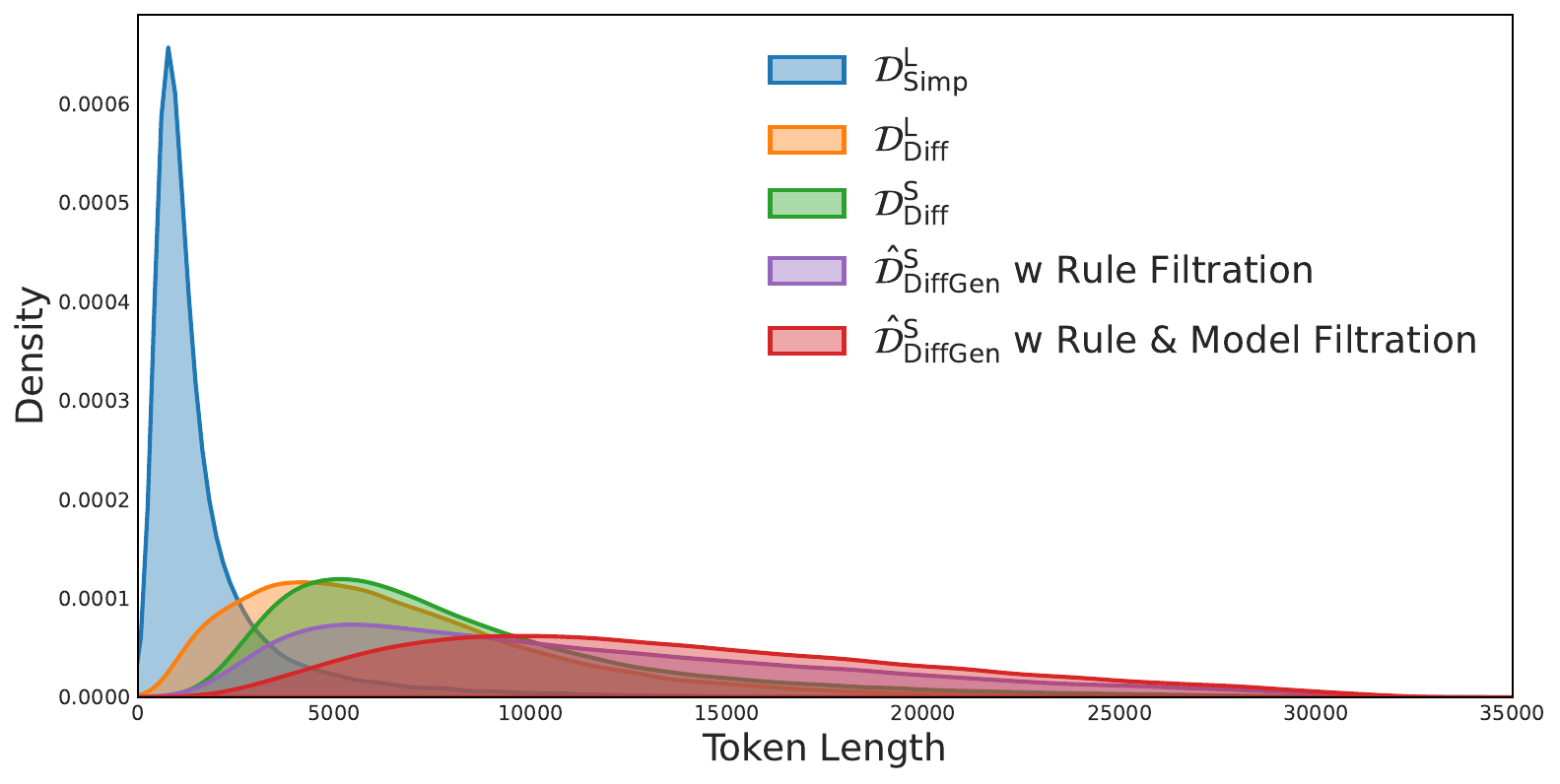}
    \caption{Distribution of solution lengths across datasets and teacher models. The superscript L denotes the use of the large-sized Qwen3-235B-A22B as the teacher model, whereas S indicates the use of the small-sized Qwen3-8B.}
    \label{fig:token_len_dist}
\end{figure}

As described in Section~\ref{sec:resp_distill_filt}, approximately 88\% of the problems generated by \gen are verified as difficult. To further investigate the characteristics of these problems, we analyze the distribution of response lengths across different datasets, namely $\mathcal{D}_{\text{Simp}}$, $\mathcal{D}_{\text{Diff}}$, and $\hat{\mathcal{D}}_{\text{DiffGen}}$, as well as across different teacher models, Qwen3-235B-A22B and Qwen3-8B (use superscript L and S to represent them, respectively). 
The results are illustrated in Figure~\ref{fig:token_len_dist}, from which several findings emerge. 
(1) Comparing the distribution of $D_{Simp}^{L}$ (blue curve) with the others, we observe that the difficulty levels identified by \textit{AdaptThink} strongly correlate with response length: simple problems exhibit a sharp density peak at very short token lengths, reflecting their requirement for only brief solution traces, while difficult problems shift the distribution toward longer token lengths, consistent with the need for more elaborate reasoning chains. 
(2) Comparing the distribution of $D_{Diff}^{L}$ and $D_{Diff}^{S}$ (orange and green curves), we find that the choice of teacher model from the same family (Qwen3) has little impact on the response length distribution for difficult problems, as both yield similar patterns.
(3) Comparing the distribution of $D_{Diff}^{L}$ and $\hat{D}_{DiffGen}^{S}$ with rule and model filtration (orange and red curves), it becomes evident that generated problems tend to induce longer responses than original difficult problems, indicating higher intrinsic complexity. 
This observation is further corroborated by downstream results in Table~\ref{tab:ablation} and~\ref{tab:teacher_model}: SFT on the 192K $D_{Diff}^{L}$ dataset yields an average performance of 56.6, whereas training on an equal amount of $\hat{D}_{DiffGen}^{S}$ with rule and model filtration achieves 58.4. 
(4) Finally, comparing the distribution of $\hat{D}_{DiffGen}^{S}$ with rule filtration and $\hat{D}_{DiffGen}^{S}$ with rule and model filtration (purple and red curves) shows that model filtration further refines the dataset by removing relatively easier problems, thereby retaining a subset of problems with greater difficulty and longer reasoning traces.

\section{Related Work}
\subsection{Mathematical Data for LRMs}

Numerous datasets have been proposed to enhance the mathematical reasoning capabilities of LRMs through SFT. Prevalent strategies~\citep{he2025deepmath,li2024numinamath,amini2019mathqa} involve \textit{selecting data from existing sources} such as textbooks, examinations and websites.
Beyond simple selection, some research focuses on data augmentation of these existing resources. \textit{Answer augmentation} methods~\citep{moshkov2025aimo,toshniwalopenmathinstruct-2,dartmath,pan2025lemma,toshniwal2024openmathinstruct,lin2025metaladder,wang2025critique} use a teacher model to synthesize novel and diverse solutions for existing problems, aiming to boost the student model's performance. These methods are often referred to as data distillation. In contrast, \textit{Question augmentation} methods~\citep{wizardmath,metamath,toshniwalopenmathinstruct-2,muggle_math,lu2024mathgenie,mitra2024orcamath,pei2025mathfusion,pan2025rest} involve synthesizing novel problems and their corresponding solutions. This method can expand topical coverage, introduce more diverse problem structures, though it requires rigorous validation to ensure the correctness of synthesized questions and solutions~\citep{cot_self_inst,li2024synthesizing}. To further enhance the diversity of synthetic data, \textit{Persona-based augmentation} technique~\citep{ge2024scaling,lambert2024tulu,li2023camel,luo2024personamath} has emerged. By incorporating role-playing into prompts, LLMs can generate diverse, role-specific mathematical problems. 
However, while some efforts have emerged to synthesize new questions, they do not explicitly control the data difficulty. Consequently, the generated problems often lack sufficient challenge for current top-tier LRMs, leading to limited improvements.

\subsection{Difficulty-aware Data Selection and Synthesis}

Data difficulty is a crucial metric for assessing data quality, significantly impacting the training effectiveness~\citep{chenadvancing}. 
Previous research has explored \textit{difficulty-aware question selection}. For example, S1~\citep{muennighoff2025s1simpletesttimescaling} and Light-R1~\citep{wen2025light} filter out simple problems that small models can easily solve, retaining difficult ones for SFT. AceReason~\citep{acereason_1} further incorporates difficulty-based filtering into RL training.
DeepMath-103K~\citep{he2025deepmath} proposes a new dataset with a higher proportion of challenging problems.
However, these methods are limited to selecting from existing data and cannot generate new, challenging examples. Furthermore, most of these techniques assess difficulty by \textit{fail rate}, a model-specific metric, which may restrict their generalizability across different models.

Another line of research focuses on \textit{synthesizing new data with varying difficulty}. 
In the mathematical domain, DART-Math~\citep{dartmath} synthesizes more solutions for difficult problems, enhancing response diversity for challenging questions.
$\text{MATH}^2$~\citep{shah2025aiassist} extracts core ``skills'' from existing math datasets and employs them as the basis for generating novel and difficult questions by prompting LLMs.
DAST~\citep{xue2025dast} proposes a difficulty-matching few-shot prompting method, presenting longer, more detailed examples for harder questions. 
ScaleQuest~\citep{ding2025unleashingllmreasoningcapability} introduces Question Preference Optimization (QPO), which optimizes generated mathematical problems based on solvability and difficulty. 
The optimized questions then serve as positive samples for preference optimization of the question generator.
MathSmith~\citep{zhan2025mathsmith} generates math problems from scratch using concept–explanation pairs, achieving superior performance on olympiad-level benchmarks.
However, most of these methods are not specifically designed for synthesizing difficult math problems. The difficulty of their generated questions remains limited, leading to restricted performance improvements.

\section{Conclusion}

In this work, we introduce \method, a simple yet effective pipeline for scaling the construction of difficult mathematical problems to enhance the complex reasoning abilities of LRMs.
By leveraging \textit{AdaptThink} as an efficient difficult problem identifier, training a dedicated generator (\gen) to produce new challenging problems, and applying both rule and model filtering to distill high-quality solutions, we construct the \method-Math dataset.
Extensive experiments demonstrate that fine-tuning on this dataset yields substantial improvements over both strong SFT- and RL-based baselines across multiple mathematical reasoning benchmarks, including AIME'24, AIME'25, HMMT-Feb'25, BRUMO'25, and MATH500. 
Moreover, we observe a clear phenomenon that augmenting training data with increasing quantities of difficult problems consistently improves performance on challenging benchmarks, underscoring the value of difficulty-aware augmentation for advancing reasoning capabilities. 
We believe that \method offers a practical and generalizable strategy for the community to strengthen LRMs, particularly in domains where complex reasoning is essential but difficult training data is scarce.

\bibliographystyle{unsrtnat}
\setcitestyle{numbers}
\bibliography{reference}

\clearpage

\end{document}